# Minimum Encoding Approaches for Predictive Modeling

**Peter Grünwald**
CWI
Dept. of Algorithms and Architectures
P.O.Box 94079
NL-1090 GB Amsterdam, The Netherlands
http://www.cwi.nl/~pdg/

**Petri Kontkanen, Petri Myllymäki,
Tomi Silander, Henry Tirri**
Complex Systems Computation Group (CoSCo)
P.O.Box 26, Department of Computer Science
FIN-00014 University of Helsinki, Finland
http://www.cs.Helsinki.FI/research/cosco/

## Abstract

We analyze differences between two information-theoretically motivated approaches to statistical inference and model selection: the *Minimum Description Length (MDL)* principle, and the *Minimum Message Length (MML)* principle. Based on this analysis, we present two revised versions of MML: a pointwise estimator which gives the MML-optimal single parameter model, and a volumewise estimator which gives the MML-optimal region in the parameter space. Our empirical results suggest that with small data sets, the MDL approach yields more accurate predictions than the MML estimators. The empirical results also demonstrate that the revised MML estimators introduced here perform better than the original MML estimator suggested by Wallace and Freeman.

## 1 INTRODUCTION

Two related but distinct approaches to statistical inference and model selection are the *Minimum Description Length (MDL)* principle (Rissanen, 1978, 1987, 1996), and the *Minimum Message Length (MML)* principle (Wallace & Boulton, 1968; Wallace & Freeman, 1987). Both approaches are based on the idea that the more we are able to *compress* a given set of data, the more we have learned about the domain the data was collected from. Nevertheless, as discussed in (Baxter & Oliver, 1994), there are subtle differences between these two approaches in both the underlying philosophy and the proposed formal criteria.

We stress that this paper concerns the general MDL *principle*, not the original *MDL model selection criterion* (Rissanen, 1978). The latter takes the same form as the Bayesian BIC criterion (Schwarz, 1978), which has led some people to believe that 'MDL=BIC' (see the discussion in (Rissanen, 1996)). The instantiation of MDL we discuss here is *not* directly related to BIC.

Recently (Rissanen, 1996), the MDL approach has been refined to incorporate effects on the description length of the data that are due to local geometrical properties of the hypothesis space. Wallace and Freeman have already taken these properties into account in their paper on MML estimators (Wallace & Freeman, 1987). It has been informally claimed by several people at several conferences that a large part of Rissanen's 1996 work is already implicit in (Wallace & Freeman, 1987). In the present paper we investigate this claim, and show that it does not hold: though superficially similar, the refinement of MDL proposed in (Rissanen, 1996) is quite different from the MML approach proposed in (Wallace & Freeman, 1987). The difference is even quite dramatic in the sense that in the MDL approach the likelihood of the data, given a model $\theta$, is multiplied by a factor correcting for the curvature of the model space near $\theta$, while in the MML approach it is *divided* by the very same factor. Our analysis of the reasons for this difference shows that there is a notable weakness in the derivation of MML estimators presented in (Wallace & Freeman, 1987). By removing this weakness, we arrive at two revised versions of MML. At this point we would like to emphasize that we are not claiming that there is anything wrong with the MML principle per se: the problem we discovered concerns only what Wallace and Freeman call 'MML estimators', which are an approximation of the theoretical MML principle for model classes with a fixed number of parameters.

In the theoretical part of the paper, in Section 2 we first briefly review the MDL and MML principles, and discuss their basic differences and similarities. In Section 3, we review in detail how the MML estimators were derived in (Wallace & Freeman, 1987), and point out an important oversight in the derivation. Based on this analysis, we present two 'revised' versions of MML: a pointwise estimator which gives the MML-optimal *single* parameter value $\theta$, and a volumewise estimator which gives the MML-optimal *region* in the parameter space. In Section 4, we show how the *two-part code MDL*, the form of MDL that most resembles MML, can be refined by the considerations presented in (Rissanen, 1996).

In Section 5 we discuss how to construct different predictive distributions based on the MML and MDL estimators considered. It turns out that the volumewise MML estimator suggested in this paper yields a predictive distribution which is closely related to the predictive distribution obtained by using Rissanen's 1996 version of MDL. Nevertheless, as the results derived are all asymptotic in nature, it is not a priori clear whether they are relevant for small sample size cases. Since one of the main points in both MDL and MML is to obtain good results for small sample sizes, this raises the interesting question of whether the original version of the MML estimator actually performs worse for small samples than either our revised MML version or the MDL approach. In Section 6 we study this question empirically, and report results demonstrating differences in performance that are consistent with the theoretical observations.

## 2 THE MML AND MDL PRINCIPLES

We assume that all data is recorded to a finite accuracy, which implies that the set $\mathcal{X}$ of all possible data values is countable. In the following we assume we are given a data sequence $\mathbf{x}^n = (\mathbf{x}_1, \ldots, \mathbf{x}_n)$ of $n$ outcomes, where each outcome $\mathbf{x}_i$ is an element of the set $\mathcal{X}$. The set of all such joint outcomes is denoted by $\mathcal{X}^n$, the $n$-fold Cartesian product of $\mathcal{X}$.

We now consider the case with a parametric family of *candidate models* $\mathcal{M} = \{f(\mathbf{x}|\theta) \mid \theta \in \Gamma\}$, where $\Gamma$ is an open bounded region of $\mathbf{R}^k$ and $k$ is a positive integer. We denote by $\mathbf{B}^*$ the set of all finite binary strings. By a (prefix) *code* $C$ we mean a one–one function from a countable set $A$ to $\mathbf{B}^*$, where the mapping is such that the *Kraft inequality* (Cover & Thomas, 1991) holds:

$$\sum_{\mathbf{x} \in A} 2^{-L_C(\mathbf{x})} \leq 1. \qquad (1)$$

Here $L_C(\mathbf{x})$ is the length (number of bits) of $C(\mathbf{x})$, the encoding of $\mathbf{x}$. The Kraft inequality allows us to interpret each probability distribution $P$ over the set $\mathcal{X}^n$ as a code $C_P$ such that for each $\mathbf{x}^n \in \mathcal{X}^n$, $L_{C_P}(\mathbf{x}^n) = -\log P(\mathbf{x}^n)$ (we assume all logarithms in this paper to be binary).

By encoding continuous outcomes to an arbitrary but finite and fixed precision, we may also regard each density function $f(\mathbf{x}|\theta)$ as a code $C_f$ with codelengths $L_{C_f}(\mathbf{x}) = -\log f(\mathbf{x}|\theta)$ (see (Rissanen, 1987) for details). We call $C_f$ the *code corresponding to* $f$. Similarly, for each code $C$ for the set $A$, we may regard $P_C$ (defined by $P_C(\mathbf{x}) = 2^{-L_C(\mathbf{x})}$) as a (possibly subadditive) probability distribution over $A$. These reinterpretations of codes as probability distributions and vice versa are basic to the minimum encoding approaches considered here: the approaches aim at constructing a code that encodes data sequences $\mathbf{x}^n$ as efficiently as possible, which corresponds to constructing a probability distribution that gives the data sequences as high a probability as possible.

A *two-part code* consists of a code $C_1$ for parameter values ('hypotheses') and a set of codes $C_{2,\theta}$ for encoding data sequences with the help of those parameter values. Since the set of parameter values $\Gamma$ is uncountable, the code $C_1$ cannot have codewords for all of them; rather, $C_1$ will be a function $C_1 : Q \to \mathbf{B}^*$ where $Q$ is some countable subset of $\Gamma$.

A data sequence $\mathbf{x}^n$ can be encoded in two steps by first encoding a parameter value $\theta \in Q$, and then encoding $\mathbf{x}^n$ by the code $C_{2,\theta}$, the code corresponding to $\theta$. As discussed above, each sequence $\mathbf{x}^n$ can be coded by using $-\log f(\mathbf{x}^n|\theta)$ bits. The two-part code $C_{1,2}$ is now defined as the code that codes each $\mathbf{x}^n$ using the $\hat{\theta} \in Q$ that is optimal for $\mathbf{x}^n$:

$$\begin{aligned} L_{C_{1,2}}(\mathbf{x}^n) &= L_{C_1}(\hat{\theta}) - \log f(\mathbf{x}^n|\hat{\theta}), \text{ where} \quad (2) \\ \hat{\theta} &= \arg\min_{\theta \in \Gamma}\{L_{C_1}(\theta) - \log f(\mathbf{x}^n|\theta)\}. \end{aligned}$$

The basic idea behind *Minimum Message Length (MML)* modeling is to find a two-part code and an associated estimator minimizing the expected message length (number of bits needed to encode the data), where the expectation is taken over the marginal distribution $r(\mathbf{x}^n)$ over the data $\mathbf{x}^n$, $r(\mathbf{x}^n) = \int_{\theta \in \Gamma} f(\mathbf{x}^n|\theta)h(\theta)d\theta$, for all $\mathbf{x}^n \in \mathcal{X}^n$. Hence any MML analysis depends on a prior distribution $h$ over the set of parameter values $\Gamma$. Interpreted in an 'orthodox' Bayesian manner, the prior $h$ represents the prior knowledge one has about the parameter values (Wallace & Freeman, 1987).

MML thus seeks to find the combination of the code $C_1$ and the estimator $\hat{\theta} : \mathcal{X}^n \to \Gamma$ minimizing the sum

$$\sum_{\mathbf{x}^n \in \mathcal{X}^n} r(\mathbf{x})[L_{C_1}(\hat{\theta}(\mathbf{x}^n)) - \log f(\mathbf{x}^n|\hat{\theta}(\mathbf{x}^n))]. \quad (3)$$

The estimator $\hat{\theta}$ that is optimal in the above sense is called the *strict MML (SMML) estimator* (Wallace & Freeman, 1987). In practice, it is very hard to find the SMML estimator — in Section 3 we will present different approximative MML estimators.

The basic principle behind *Minimum Description Length (MDL)* modeling is to find a code (not necessarily an estimator) that minimizes the code length over all data sequences which can be well modeled by $\mathcal{M}$. Here a data sequence is 'well-modeled by $\mathcal{M}$' means that there is a model $\theta$ in $\mathcal{M}$ which gives a good fit to the data. In other words, if we let $\tilde{\theta}(\mathbf{x}^n)$ denote the maximum likelihood estimator (MLE) of the data $\mathbf{x}^n$, then '$\mathbf{x}^n$ is well modeled by $\mathcal{M}$' means that $f(\mathbf{x}^n|\tilde{\theta}(\mathbf{x}^n))$ is high. The *stochastic complexity* of a data sequence $\mathbf{x}^n$, relative to a family of models $\mathcal{M}$, is the code length of $\mathbf{x}^n$ when it is encoded using the most efficient code obtainable with the help of the family $\mathcal{M}$. There exist several alternative ways for defining the stochastic complexity measure and the MDL principle explicitly. We return to this issue in Section 4.

A crucial difference between the MML and MDL principles is that the former is based on finding a code minimizing *expected* codelengths, while the latter is based on finding a code that yields short codelengths for *all* datasets that are well-modeled by $\mathcal{M}$. Another important difference is that the goal of the MML approach is to find an efficient code together with the associated estimator, while the MDL definition is in general only concerned with the codes. What is more, the MML approach uses for this purpose always two-part codes; for MDL there are several options, of which the so called two-part code MDL is only one special case.

It should also be noted that while MML is Bayesian in the sense that the approach is dependent on a subjective prior provided by an external observer, the MDL principle does not depend on any specific prior distribution. To be sure, priors do arise in MDL modeling, but they are merely used as technical tools and not as representing prior knowledge about the problem at hand.

## 3 MML ESTIMATORS

### 3.1 MMLWF ESTIMATOR

We now consider the original derivation presented in (Wallace & Freeman, 1987), and call the resulting MML estimator the *MMLWF estimator*. We concentrate first on the case of a model class $\mathcal{M}$ containing models depending on a single parameter (hence $\Gamma \subset \mathbf{R}^1$). Rather than trying to find the code optimizing (3), we now consider the following problem: given an observed data sequence $\mathbf{x}^n$, we are asked to choose an estimate $\theta' \in \Gamma$ together with a precision quantum $d$, so that if $\mathbf{x}^n$ is encoded by first stating $\theta'$ with precision $d$ and then stating $\mathbf{x}^n$ using the code corresponding to the stated estimate, then the length of the encoding is minimized. This means that the estimate $\theta'$ is coded using only a limited number of binary places; in other words, a truncated value $\hat{\theta}$ is obtained from $\theta'$ by selecting a value from a quantized scale in which adjacent values differ by $d$: $|\hat{\theta} - \theta'| \le d/2$. Coding $\mathbf{x}^n$ with the obtained estimate requires $L(\mathbf{x}^n) = -\log f(\mathbf{x}^n|\hat{\theta})$ bits.

Note that while in the SMML setup the goal was to minimize (3), now we only ask for a 'target' estimate $\theta'$ together with a precision quantum $d$. Consequently, in contrast to SMML, here we do not require the detailed coding of the actually used estimate $\hat{\theta}$ to be specified. This makes the approach feasible; the price we pay is that the exact effect of encoding the data using the quantized value $\hat{\theta}$ instead of $\theta'$ cannot be predicted, and the code length can be minimized only in expectation.

We assume that the quantization has the following effects: $E(\theta' - \hat{\theta}) = 0$ (unbiasedness), $E[(\theta' - \hat{\theta})^2] = d^2/12$ (as for a uniform distribution). The prior probability that $\theta$ lies within $\pm d/2$ of a quantized value $\hat{\theta}$ is approximately $d \cdot h(\hat{\theta})$. Let $C_{\text{WF}} : Q \to \mathbf{B}^*$ be the code with lengths corresponding to this probability. Encoding the estimates $\hat{\theta}$ using $C_{\text{WF}}$, the expected length of the first part stating $\theta'$ to precision $d$ is $-\log dh(\theta')$. Using the expectation of the effects of this quantization, and approximating the length of the second part by the Taylor expansion (to second order), we get a code length of

$$-\log dh(\theta') - \log f(\mathbf{x}^n|\theta') + \frac{d^2}{24} I(\mathbf{x}^n; \theta'), \quad (4)$$

where $I(\mathbf{x}^n; \theta')$ is short for $-\frac{\partial^2}{\partial \theta^2} \log f(\mathbf{x}^n|\theta')$. The expected codelength (4) is minimized by choosing $d^2 = 12/I(\mathbf{x}^n; \theta')$. Substituting this optimal precision, we get the expected code length to be

$$-\log h(\theta') + \frac{1}{2} \log \frac{I(\mathbf{x}^n, \theta')}{12} - \log f(\mathbf{x}^n|\theta') + \frac{1}{2}. \quad (5)$$

The value $\theta'$ which minimizes this is called the *MML estimate*.

It is clear that in order to decode a two-part message as used here, one must first decode the parameter value $\hat{\theta}$. For this, one must know the precision $d$ that was used to encode $\hat{\theta}$. Since the optimal precision depends on $\mathbf{x}^n$, it is not constant and hence it seems that it must be made part of the code too. However, in (Wallace & Freeman, 1987) it was shown that the minimum of the expected message length reached for the optimal precision $d^2 = 12/I(\mathbf{x}^n, \theta')$ is very broad with respect to $d$. This implies that using a quantum $d$ based on the *expectation* of $I(\mathbf{x}^n, \theta')$, rather than $I(\mathbf{x}^n, \theta')$ itself, will be reasonably efficient for most data values. Hence we can use $d^2 = 12/I_n(\theta')$, where $I_n(\theta') = -E_\Theta(\partial^2 \log f(\mathbf{x}^n|\theta)/\partial\theta^2)$ which coincides with the *Fisher (expected) information* for $n$ observations (Berger, 1985).

The advantage of using $I_n(\theta')$ is that now the optimal $d$ is independent of the observed data and becomes a function of $\theta'$ only. This means that there is only one set of possible truncated estimates which can be constructed without reference to the data. We can thus construct a code for the estimate which does not need a precision preamble (for more details we refer to (Wallace & Freeman, 1987)). From (5) we see that the final definition of the MMLWF estimator becomes

$$\theta' = \arg\max_{\theta \in \Gamma} \frac{f(\mathbf{x}^n|\theta)h(\theta)}{\sqrt{I_n(\theta)}}. \qquad (6)$$

## 3.2 MMLP ESTIMATOR

Let us define $d: \Gamma \to \mathbf{R}$ as a function which gives for each value $\theta'$ the corresponding optimal precision quantum $d(\theta')$. Using this notation, and substituting $I_n(\theta')$ for $I(\mathbf{x}^n; \theta')$ (as prescribed at the end of the previous section), the expected total codelength $L(\mathbf{x}^n, \theta')$ given in (4) can be rewritten as

$$-\log d(\theta') - \log h(\theta') - \log f(\mathbf{x}^n|\theta') + \frac{d(\theta')^2}{24} I_n(\theta'). \qquad (7)$$

We now make two assumptions. First, we assume that the value $\theta'$ which minimizes the expected code length may in principle lie anywhere in the interior of the parameter space. Second, we assume that the number of different possible truncated parameter values is finite, say $N$. Consequently, we can write $\Gamma = Q_N = \{\hat\theta_1, \ldots, \hat\theta_N\}$. Both assumptions are quite reasonable. For example, the first assumption follows from the requirement that $\theta'$ should be consistent, together with the (much stronger) assumption that there exists a true value $\theta$ according to which the data is actually drawn, and which may lie at any point in the interior of $\Gamma$. The MML estimator — both the MMLWF version (6) and our 'corrected' version below — is indeed consistent for all combinations of priors and model classes $\mathcal{M}$ for which the Bayesian maximum posterior (MAP) estimator is asymptotically normally distributed around its true parameter value. This is true because $I_n(\theta)$ will be constant in the neighborhood of that value. The second assumption is reasonable as long as we allow $N$ to grow with the number of observations $n$, as we indeed do. We only assume that for each fixed $n$ there is a finite number of candidates $N$.

From (7) we see that the expected effect of the quantization depends in two ways on the region in the parameter space where $\theta'$ lies: first, through the term $-\log d(\theta')$ and second through the term proportional to $d(\theta')^2 I_n(\theta')$. Now the point that (in our view) has been overlooked by Wallace and Freeman is that we can eliminate the effect of the first term altogether without influencing any of the other terms. To see this, let us first consider the special case that $h$ is the uniform prior ($h(\theta) = c$ for some constant $c$). We will now code parameter values using the *uniform code* $C_{\text{UNI}}$ rather than the code $C_{\text{WF}}$. The uniform code is simply the code that codes each element of $Q_N$ using the same number of bits $\log N$. It follows directly from the Kraft inequality (1) that for every other code $C': Q_N \to \mathbf{B}^*$, we have $L_{C'}(\hat\theta_i) > L_{C_{\text{UNI}}}(\hat\theta_i)$ for at least one $\hat\theta_i \in Q_N$. We say that $C_{\text{UNI}}$ has the *optimal worst-case codelength* (here 'optimal' is used in the sense of 'shortest').

Using $C_{\text{UNI}}$ instead of $C_{\text{WF}}$, the codelength to encode $\hat\theta$ becomes $\log N$ instead of $-\log d \cdot c$, and (7) becomes

$$\log N - \log f(\mathbf{x}^n|\theta') + \frac{d(\theta')^2}{24} I_n(\theta'). \qquad (8)$$

For some $\theta'$ this will yield shorter codelengths than (7) while for others it will yield larger ones. However, using our assumption that $\theta'$ may in principle lie everywhere in the interior of $\Gamma$, we should take a worst-case point of view[1]: for the worst-case $\theta'$, the expected codelength (8) is clearly smaller than the expected length (7). Consequently, because of the worst-case optimality of $C_{\text{UNI}}$, $C_{\text{UNI}}$ should be preferred over $C_{\text{WF}}$, and indeed over any other possible code for the set $Q_N$.

We next consider the general case for arbitrary priors $h$. In this case, instead of using the code $C_{\text{UNI}}$, we may use the modified uniform code $C_{\text{UNI},h}$ which corrects for our prior 'beliefs' that are encoded by $h$: $L_{C_{\text{UNI},h}}(\hat\theta) = \log N - \log h(\theta')$. We may interpret this code as a prior distribution that transforms the density $h$ to a probability distribution $H$ on $Q_N$. This follows from the fact that $L_{C_{\text{UNI},h}}(\hat\theta) = -\log H(\hat\theta)$, where

$$H(\hat\theta) = \frac{h(\hat\theta)}{N}. \qquad (9)$$

We can now motivate the use of $C_{\text{UNI},h}$ also from a Bayesian point of view: $H(\hat\theta)$ makes the prior probability mass of each truncated parameter value $\hat\theta$ pro-portional to its prior density $h(\hat{\theta})$, and as such it remains faithful to the prior density. This is not the case for the code $C_{\text{WF}}$.

---

[1] From a Bayesian point of view, one may argue that we should not take such a worst-case viewpoint since some regions for $\theta'$ may be a priori more likely than others. A Bayesian justification for the choice of $C_{\text{UNI}}$ will be given after Equation (9).

Using $C_{\text{UNI},h}$, codelength (8) becomes

$$\log N - \log h(\theta') - \log f(\mathbf{x}^n|\theta') + \frac{d(\theta')^2}{24} I_n(\theta'). \quad (10)$$

We see from this that using $C_{\text{UNI},h}$ for encoding the parameter values is optimal independently of the way $d(\theta')$ is instantiated. This means that we should base our two-part codes on (10) rather than (7), and furthermore instantiate (10) by using the function $d(\theta')$ that gives shortest expected codelengths. Taking once again the worst-case point of view, $\theta'$ may lie anywhere in $\Gamma$, so the optimal $d(\theta')$ becomes the function that minimizes the maximum value of the last term in (10):

$$d = \arg\min_d \max_{\theta' \in \Gamma} d(\theta')^2 I_n(\theta'), \quad (11)$$

which is clearly[2] attained for $d(\theta')^2 \propto I_n(\theta')^{-1}$. By substituting this optimal $d$ back into (10), we obtain

$$L_{\text{OPT}} \approx -\log h(\theta') - \log f(\mathbf{x}^n|\theta') + \log N + K, \quad (12)$$

where $K$ depends only on $N$ and $n$, and not on $\theta$. The $\theta'$ which minimizes this, however, is simply the standard Bayes posterior mode! Thus, interestingly, we find that in our 'corrected' derivation of MML estimators, the optimal MML estimate is just the (Bayesian) MAP estimate. On the other hand, the optimal precision $d(\theta')$ at point $\theta'$ remains inversely proportional to $\sqrt{I_n(\theta')}$, just like in the derivation of the MMLWF estimator.

### 3.3 MMLV ESTIMATOR

Using $C_{\text{UNI},h}$, we can code the data by first stating a $\hat{\theta}_i \in Q_N$ using $\log N - \log h(\hat{\theta}_i)$ bits, and then stating $\mathbf{x}^n$ using $-\log f(\mathbf{x}^n|\hat{\theta}_i)$ bits. Using the correspondence between codes and probability distributions, this can equivalently be recast as determining the posterior probability of $\hat{\theta}_i$ given data $\mathbf{x}^n$, using the discrete prior $H(\hat{\theta}_i)$ as given by (9). We denote this probability by $P$:

$$P(\hat{\theta}_i|\mathbf{x}^n) \propto f(\mathbf{x}^n|\hat{\theta}_i) H(\hat{\theta}_i). \quad (13)$$

In this probabilistic formulation, (12) tells us that the *single* value $\theta'$ which maximizes the expected value of $P(\hat{\theta}_i|\mathbf{x}^n)$ (where $\hat{\theta}_i$ is the truncated version of $\theta'$), is given by the MAP estimate. However, it tells us nothing about the width of the minimum attained. Indeed, as we shall see, it may be extremely narrow. It may therefore be more interesting to choose a small (but non-zero) width $w$, and look for the interval in $\Gamma$ of width $w$ with the maximal posterior probability mass, or equivalently, the shortest codelength according to (13).

To obtain this interval, let us adapt the line of reasoning used in (Rissanen, 1996), and look at the MML two-part code in another manner. We partition the parameter space $\Gamma$ into a set of adjacent regions $R_1, \ldots, R_M$, each of width $w$, where $w$ is such that $M \ll N$. Let us now determine the region $R_i$ with maximum posterior probability mass $P(R_i|\mathbf{x}^n)$. We first associate with each region $R_i$ the element $\theta_i$ that lies in the center of $R_i$, so $R_i = [\theta_i - w/2, \theta_i + w/2]$. We can now extend the density $f(\mathbf{x}^n|\theta_i)$, determined by a single value $\theta_i$, to a probability determined by a region in the parameter space $R_i$ in the obvious way by defining $f(\mathbf{x}^n|R_i) = \int_{R_i} f(\mathbf{x}^n|\theta)\pi(\theta)d\theta$, where $\pi$ is an arbitrary proper prior with support $R_i$. In the limit (for small $R_i$), we have $f(\mathbf{x}^n|R_i) = f(\mathbf{x}^n|\theta_i)$. This implies

$$P(R_i|\mathbf{x}^n) \propto f(\mathbf{x}^n|R_i) H(R_i), \quad (14)$$

where $H$ is as given in (9). Marginalizing over the values $\hat{\theta} \in Q_N$ contained in $R_i$, we find that

$$H(R_i) = \sum_{\hat{\theta} \in R_i \cap Q_N} \frac{h(\hat{\theta})}{N}. \quad (15)$$

In the limit for large $N$, we may select the regions $R_i$ small enough so that for all regions $R_i$, we can regard $f(\mathbf{x}^n|\theta)$, $h(\theta)$ and $\sqrt{I_n(\theta)}$ as approximately constant for all $\theta$ within a single interval $R_i$.

Let $|R_i \cap Q_N|$ denote the number of parameter values in $R_i$ that are represented in the set of encodeable parameters $Q_N$. For the optimal precision quantum $d(\theta)$, we have $d^2(\theta) \propto I_n(\theta)^{-1}$. It follows that in the limit for large $N$, the density of parameter values $\hat{\theta}$ in region $R_i$ will be proportional to $\sqrt{I_n(\theta_i)}$:

$$|R_i \cap Q_N| = w\sqrt{I_n(\theta_i)} \cdot c, \quad (16)$$

where $c$ is a constant not depending on $i$. Since in the limit for large $N$, we may pick $w$ as small as we please, we can assume that $h(\hat{\theta}) \approx h(\theta_i)$ for all $\hat{\theta} \in R_i$. We then have from (15) and (16) that $H(R_i) \propto w\sqrt{I_n(\theta_i)} h(\theta_i)$. We now conclude from (14), together with the fact that $w$ does not depend on $i$, that $P(R_i|\mathbf{x}^n) \propto f(\mathbf{x}^n|\theta_i) h(\theta_i) \sqrt{I_n(\theta_i)}$. The region $R_i$ which maximizes this is the most probable posterior region if the prior $H$ is set according to (9). Assuming that both $h$ and $I$ are continuous functions of $\theta$, the $\theta \in \Gamma$ yielding the shortest expected codelength *in its neighborhood* is thus given by

$$\theta'' = \arg\max_{\theta \in \Gamma} f(\mathbf{x}^n|\theta) h(\theta) \sqrt{I_n(\theta)}. \quad (17)$$

---

[2] Note that choosing $d(\theta') = 0$ everywhere is not an option since we assume that there exist only $N$ parameter values.

We see that in our 'corrected' MML derivation, choosing the parameter value with the highest probability content in its neighborhood (17) gives us an estimate which maximizes the Bayesian posterior *times* $\sqrt{I_n(\theta)}$. This is in sharp contrast with the original MML estimate (6) which maximizes the Bayesian posterior *divided* by $\sqrt{I_n(\theta)}$!

For simplicity, our derivations above have been only for the case where $\Gamma \subset \mathbf{R}^1$. The generalization to the multiple-parameter case is completely straightforward: it suffices to replace the intervals $R_i$ of width $w$ by *rectangles* $R_i$ of *volume* $w$. In this case, the square root of the Fisher information $\sqrt{I_n(\theta)}$ becomes $\sqrt{|I_n(\theta)|}$, the square root of the determinant of the Fisher information matrix given by

$$[I_n(\theta)]_{i,j} = -E_\Theta \left[ \frac{\partial^2 \log f(\mathbf{x}^n|\theta)}{\partial \theta^i \partial \theta^j} \right]. \quad (18)$$

## 4  MDL ESTIMATORS

The MDL principle aims at finding an efficient coding for all data sequences $\mathbf{x}^n$. One obvious possibility for this is to use a two-part code as with the MML approach; the resulting two-part code MDL estimator is discussed in (Rissanen, 1978, 1989). However, it is relatively easy to see that the two-part code (2) is redundant: every data sequence $\mathbf{x}^n$ can be encoded using every $\hat{\theta} \in Q$ for which $f(\mathbf{x}^n|\hat{\theta}) > 0$. Though $C_{1,2}$ will always use the particular $\hat{\theta} \in Q$ for which the total description length is minimized, codewords are 'reserved' for many other ways of encoding $\mathbf{x}^n$. Until recently, it has not been clear how to remove this redundancy in a principled manner. In (Rissanen, 1996), this problem was finally solved.

Following Rissanen, for simplicity we assume that $Q$ contains a finite number of parameters. As in Section 3, we can thus write $Q = Q_N = \{\hat{\theta}_1, \ldots \hat{\theta}_N\}$. We also assume uniform codelengths for the models: $L_{C_1}(\hat{\theta}_i) = \log N$ for all $\hat{\theta}_i \in Q_N$. Rissanen observed that a decoder, after having decoded $\hat{\theta}_i$, already knows something about the data $\mathbf{x}^n$ whose description will follow. Namely, he knows that $\mathbf{x}^n$ must be a member of a subset $\mathcal{D}_i$ of the set of all possible data $\mathcal{X}^n$, where $\mathcal{D}_i$ is the set of all data $\mathbf{x}^n$ for which $\hat{\theta}_i$ gives the shortest two-part codelength. The reason for the decoder knowing that $\mathbf{x}^n \in \mathcal{D}_i$ after decoding $\hat{\theta}_i$ is the following: looking at equation (2), we see that if $\mathbf{x}^n \notin \mathcal{D}_i$, then the decoder would not have decoded $\hat{\theta}_i$, but rather some other $\hat{\theta}_j \neq \hat{\theta}_i$. This fact can be exploited to change the code $C_{2,\theta}$ that was used in the original two-part code (2), to a code $C'_{2,\theta}$ with strictly shorter lengths. Using $C'_{2,\theta}$, we code $\mathbf{x}^n$ not by the code corresponding to probability distribution $f(\mathbf{x}^n|\hat{\theta}_i)$, but rather by the code based on the normalized probability distribution

$$f(\mathbf{x}^n|\mathcal{M}) = \frac{f(\mathbf{x}^n|\hat{\theta}_i)}{\sum_{\mathbf{x}^n \in \mathcal{D}_i} f(\mathbf{x}^n|\hat{\theta}_i)}.$$

In this case the total description length becomes $\log N - \log f(\mathbf{x}^n|\hat{\theta}_i) + \log \sum_{\mathbf{x}^n \in \mathcal{D}_i} f(\mathbf{x}^n|\hat{\theta}_i)$ bits rather than just $\log N - \log f(\mathbf{x}^n|\hat{\theta}_i)$ bits. In general, $\sum_{\mathbf{x}^n \in \mathcal{D}_i} f(\mathbf{x}^n|\hat{\theta}_i) < 1$, which means that the revised two-part code has a strictly shorter codelength than the original one.

It was shown in (Rissanen, 1996) that the normalization trick described above can be optimally exploited (for large $N$) if, for every $\theta \in \Gamma$, the density of parameter values in $Q_N$ in the neighborhood of $\theta$ is proportional to $\sqrt{|I(\theta)|}$, where $|I(\theta)|$ is the determinant of the Fisher information matrix (18). This means that either the spacing between any two adjacent values $\hat{\theta}_i$ and $\hat{\theta}_{i+1}$ in $Q_N$ should be made proportional to $1/\sqrt{|I(\theta)|}$, or the code giving codelengths $\log N$ to every $\hat{\theta}_i \in Q_N$ should be changed. Rissanen chooses the second option, but explicitly mentions that the first one is possible too (Rissanen, 1996, page 43).

We now see the reason for the confusion mentioned in the introduction: although the optimal width between adjacent parameter values as determined in (Wallace & Freeman, 1987) is *also* proportional to $1/\sqrt{|I(\theta)|}$, this same width was chosen for a very different reason. What is more, as we shall see in Sections 5 and 6, making predictions of future data on the basis of MMLWF-estimator can be quite different from making predictions on the basis of Rissanen's 1996 MDL estimator.

## 5  MINIMUM ENCODING PREDICTIVE INFERENCE WITH BAYESIAN NETWORKS

In the context of his 1996 paper, Rissanen was not interested in obtaining a single optimal model for the observation sequence $\mathbf{x}^n$, but rather obtaining an optimal predictive distribution for the prediction of future data. Using similar arguments as we employed in Section 3.3, and suitable regularity conditions on the class of models $\mathcal{M}$, Rissanen arrives at the following: the predictive distribution for predicting $\mathbf{x}_{i+1}$ on the basis of $\mathbf{x}^i = (\mathbf{x}_1, \ldots, \mathbf{x}_i)$ is

$$f(\mathbf{x}_{i+1}|\mathbf{x}^i) \propto \int f(\mathbf{x}^{i+1}|\theta) P(\theta) d\theta, \quad (19)$$

where the prior distribution $P(\theta)$ is chosen to be the so-called Jeffrey's prior $\pi(\theta)$ (Berger, 1985),

$$\pi(\theta) \propto \sqrt{|I(\theta)|}. \quad (20)$$

It is now interesting to see that our revised volumewise MML estimator $\theta''$ leads to the following predictive distribution:

$$f(\mathbf{x}_{i+1}|\mathbf{x}^i) = f(\mathbf{x}_{i+1}|\theta(\mathbf{x}^i)), \quad (21)$$

where $\theta(\mathbf{x}^i)$ is set equal to the maximum posterior probability (MAP) values $\theta''(\mathbf{x}^i)$ given by Eq. (17). We see that if we take the 'subjective' prior $h$ to be uniform, then our revised MML prediction becomes equivalent to MAP model prediction using Jeffrey's prior, while Rissanen's predictive distribution is equivalent to the Bayesian *marginal* distribution based on Jeffrey's prior. In contrast to this, our revised pointwise MML estimator would in this case lead to a predictive distribution where $\theta$ would represent the parameter values maximizing the posterior probability of the parameters with uniform prior distribution. What is more, the original MMLWF estimator uses a model $\theta'(\mathbf{x}^i)$ determined by (6), which, in the case of a uniform prior $h$, would be equivalent to MAP prediction using the inverse of Jeffrey's prior.

For most regular model classes, the predictions made using the MAP approach (21) and those based on marginal likelihood formula (19) will converge to the same values as the sample size grows to infinity. This happens independently of the specific prior being used. This implies that for large sample sizes, all the predictive methods discussed here will give approximately the same results. Consequently, the differences between the methods become relevant only for small sample sizes. Unfortunately, since both Rissanen's 1996 and our theoretical results are asymptotic in nature, they do not say too much about this situation. It is therefore an interesting empirical question, whether either Rissanen's MDL approach or our revised MML estimators lead to a more accurate predictive distribution than the MMLWF estimator in cases where only a limited amount of data is available. In the next section, we study the predictive performance of the different predictive distributions empirically by using small, real-world datasets. For being able to perform these experiments, we now instantiate the above listed different predictive distributions for a model family of practical importance, the family of *Bayesian networks* (see, e.g., (Heckerman, 1996)).

A Bayesian network is a representation of a probability distribution over a set of (in our case) discrete variables, consisting of an acyclic directed graph, where the nodes correspond to domain variables $X_1, \ldots, X_m$. Each network topology defines a set of independence assumptions which allow the joint probability distribution for a data vector $\mathbf{x}$ to be written as a product of simple conditional probabilities, $P(X_1 = x_1, \ldots, X_m = x_m) = \prod_{i=1}^{m} P(X_i = x_i | \mathrm{pa}_i = q_i)$, where $q_i$ denotes a configuration of (the values of) the parents of variable $X_i$. Consequently, in the Bayesian network model family, a distribution $P(\mathbf{x}|\Theta)$ is uniquely determined by fixing the values of the parameters $\Theta = (\theta^1, \ldots, \theta^m)$ where $\theta^i = (\theta^i_{11}, \ldots, \theta^i_{1n_i}, \ldots, \theta^i_{c_i 1}, \ldots, \theta^i_{c_i n_i})$, $n_i$ is the number of values of $X_i$, $c_i$ is the number of configurations of $\mathrm{pa}_i$, and $\theta^i_{q_i x_i}$ denotes the probability $P(X_i = x_i | \mathrm{pa}_i = q_i)$.

As demonstrated in (Cooper & Herskovits, 1992; Heckerman et al., 1995), with certain technical assumptions (Multinomial-Dirichlet model, i.i.d. data, parameter independence), the mode of the posterior parameter distribution $P(\Theta|\mathbf{x}^n)$ is obtained by setting

$$\theta^i_{q_i x_i} = \frac{f^i_{q_i x_i} + \mu^i_{q_i x_i} - 1}{\sum_{l=1}^{n_i}\left(f^i_{q_i l} + \mu^i_{q_i l}\right) - n_i}, \quad (22)$$

where $\mu^i_{q_i x_i}$ denotes the hyperparameter corresponding to parameter $\theta^i_{q_i x_i}$, and $f^i_{q_i x_i}$ are the sufficient statistics of the training data $\mathbf{x}^n$: $f^i_{q_i x_i}$ is the number of data vectors in $\mathbf{x}^n$ where variable $X_i$ has value $x_i$ and the parents of $X_i$ have configuration $q_i$.

Varying the hyperparameters $\mu^i_{q_i x_i}$ corresponds to using different prior distributions $P(\Theta)$, which furthermore lead to different predictive distributions. All three different MML estimators discussed earlier lead to the same predictive distribution form (21), where the parameters $\Theta$ are set to their MAP values (22), but the methods differ in the way they define the prior distribution $P(\Theta)$ to be used. As we saw earlier, the Wallace and Freeman MML estimator described in Section 3.1 suggests using the prior $P(\Theta) = h(\Theta)/\pi(\Theta)$, where $h(\Theta)$ is a subjective prior provided by the user, and $\pi(\Theta)$ denotes the Jeffrey's prior defined by (20). Our revised pointwise MML estimator described in Section 3.2 leads to using the subjective prior $h(\Theta)$ as the prior $P(\Theta)$. The revised volumewise MML estimator described in Section 3.2 suggests that the prior should be defined by $P(\Theta) = h(\Theta)\pi(\Theta)$.

As shown in (Kontkanen et al., 1998), the Jeffrey's prior distribution $\pi(\Theta)$ can in the above Bayesian network model family case be computed by

$$\pi(\Theta) \propto \prod_{i=1}^{m}\prod_{q_i=1}^{c_i}(P^i_{q_i})^{\frac{n_i-1}{2}}\prod_{l=1}^{n_i}(\theta^i_{q_i l})^{-\frac{1}{2}}, \quad (23)$$

where $P^i_{q_i}$ stands for the probability $P(\mathrm{pa}_i = q_i|X_i = x_i)$.

For determining the MDL predictive distribution, we need to be able to compute the predictive distribution (19). As shown in (Cooper & Herskovits, 1992; Heckerman et al., 1995), this integral can be computed by using the the MAP predictive distribution (21) with a single model $\bar{\Theta}$, where instead of using the maximum probability values given by (22) as above, $\bar{\Theta}$ is obtained by setting each parameter to its *expected* value:

$$\bar{\theta}^i_{q_i x_i} = \frac{f^i_{q_i x_i} + \mu^i_{q_i x_i}}{\sum_{l=1}^{n_i} \left( f^i_{q_i l} + \mu^i_{q_i l} \right)}. \quad (24)$$

As discussed earlier, the hyperparameters should in this case be such that the resulting prior distribution becomes the Jeffrey's prior (23).

## 6  EMPIRICAL RESULTS

For comparing empirically the four predictive distributions discussed in the previous section, we used the following six public domain classification datasets from the UCI data repository[3]: Australian (AU), Diabetes (DI), Glass (GL), Heart disease (HD), Iris (IR) and Lymphography (LY). For avoiding the computationally intensive tasks related to the problem of searching the model structure space, we fixed the model structure in this experimental setup to the structurally simple Naive Bayes model, where the variables in $X_1, \ldots, X_{m-1}$ are assumed to be independent given the value of class variable $X_m$. Consequently, we can regard the Naive Bayes model as a simple tree-structured Bayesian network, where the classification variable $X_m$ forms the root of the tree, and the other variables are represented by the leaves. Despite of its structural simplicity, this model has been demonstrated to perform well when compared to more complex models (Friedman et al., 1997; Kontkanen et al., 1997).

In the situation where only a limited amount of training data is available, using the MML predictive distributions may be technically difficult if the subjective prior $h$ is such that the corresponding hyperparameter values are very small. The reason for this is that in some cases, the expressions that are maximized in equations (6) and (17) have no maximum. Taking the supremum instead of the maximum does not help, as there are usually several different suprema (at the boundaries of the parameter space), which give rise to completely different predictions. For this reason, in this set of experiments we determined the subjective prior $h$ by using the *equivalent sample size (ESS)* priors, which have a clear interpretation from a subjective Bayesian point of view (Heckerman, 1996). Experiments with different ESS subjective priors seemed to produce similar results. In the experiments reported here, the equivalent sample sizes where chosen to be the smallest possible numbers with which the above mentioned technical difficulty did not occur.

---

[3] "http://www.ics.uci.edu/~mlearn/".

In the first set of experiments, we computed the cross-validated 0/1-scores for each of the four methods by using 5-fold crossvalidation (following the testing scheme used in (Friedman et al., 1997)). The 0/1-score is computed by first determining the class $k$ for which the predictive probability is maximized (over all the possible values of the class variable $X_m$), and the 0/1-score is then defined to be 1, if the actual outcome indeed was $k$, otherwise 0. However, as the results appeared to be strongly dependent on the way the data was partitioned in the 5 folds to be used, we repeated the whole crossvalidation cycle 10000 times with different, randomly chosen partitionings of data. The results are given in Table 1.

We can now observe that, first of all, with respect to the 0/1-score, there seems to be no clear winner between the different predictive distributions used, and the differences between the results are usually small. Secondly, it should be noted that the results vary a great deal with different partitionings of the data. As the corresponding results reported in the literature are usually obtained by using only one crossvalidation cycle (a single partitioning of data), evaluating the relevance of the earlier crossvalidation results is troublesome, and hence comparing these results to the earlier results is problematic. Nevertheless, as in some cases even the average result (not to mention the maximum) of the 10000 runs reported here is better than the corresponding single run results reported in the literature, it is evident that the minimum encoding approaches perform very well.

In the second set of experiments, instead of predicting only the value of the class variable $X_m$, we used the predictive distributions for computing the joint probability for the unseen testing vectors as a whole. In this case the accuracy of the methods was measured by using the log-score, which is defined as the negation of the logarithm of the probability given to the unseen vector to be predicted.

To prevent the large fluctuation in the results, we used in this experiment the leave-one-out form of crossvalidation, where the task is at each stage to predict one data vector, given all the others. The results of this experiment can be found in Table 2. From these results we can now see that the MDL approach produced consistently the best score, and of the MML estimators considered here, the MMLV estimator was more accurate that the MMLP estimator, which performed better than MMLWF (with the exception of the Lymphography database).

To study the small sample behavior of the methods in more detail, we rerun the leave-one-out crossvalidation experiments, but used at each stage only $s$ (randomly chosen) vectors of the available $n - 1$ vectors for producing the predictive distribution, where $s$ varied between 1 and $n - 1$. In this case, the results obtained are quite similar with all six datasets: the results with $s = 0.1n$ can be found in Table 2. As an illustrative example of the typical behavior of the methods with varying amount of training data, in Figure 1 we plot the results in the Heart disease dataset case as a function of $s$. In this figure, the log-scores are scaled with respect to the score produced by the MMLWF method so that the MMLWF method gets always a score 0, and a positive score means that the actual log-score was better than the MMLWF log-score by the corresponding amount. From Figure 1 we now see two interesting things: firstly, the different predictive distributions seem to converge with increasing amount of training data, as was expected from the discussion in Section 5. Secondly, the relative differences between the methods seem to grow when the amount of available data is decreased. The corresponding figures with the other five datasets show similar behavior. This suggests that the differences between the various approaches presented here may be practically significant in cases with small amount of data.

Table 1: Classification 0/1-scores with 10000 independent 5-fold crossvalidation runs.

|    | MIN MMLWF | MIN MMLP | MIN MMLV | MIN MDL | MEAN MMLWF | MEAN MMLP | MEAN MMLV | MEAN MDL | MAX MMLWF | MAX MMLP | MAX MMLV | MAX MDL |
|----|------|------|------|------|------|------|------|------|------|------|------|------|
| AU | 83.5 | **83.6** | 83.5 | **83.6** | **84.9** | **84.9** | 84.8 | **84.9** | **86.2** | 86.1 | 86.1 | **86.2** |
| DI | 73.4 | **73.6** | 73.3 | 73.2 | **75.5** | 75.4 | 75.4 | 75.3 | 77.1 | 77.2 | 77.2 | **77.3** |
| GL | 56.5 | 56.5 | 56.1 | **58.4** | 62.6 | 62.6 | 62.7 | **64.9** | 67.3 | 68.2 | 67.3 | **70.1** |
| HD | **81.9** | **81.9** | **81.9** | 81.1 | **84.5** | 84.4 | 84.4 | 84.1 | **87.0** | **87.0** | **87.0** | **87.0** |
| IR | **93.3** | 92.7 | 92.7 | 92.0 | **94.4** | 94.3 | 94.3 | **94.4** | 96.0 | 96.0 | **96.7** | **96.7** |
| LY | **79.1** | 78.4 | 78.4 | **79.1** | **84.2** | 84.0 | 83.6 | 83.9 | **88.5** | 87.8 | **88.5** | **88.5** |

Table 2: Leave-one-out crossvalidated log-scores in the joint probability estimation task.

|    | 10% training data MMLWF | 10% training data MMLP | 10% training data MMLV | 10% training data MDL | 100% training data MMLWF | 100% training data MMLP | 100% training data MMLV | 100% training data MDL |
|----|-------|-------|-------|-------|-------|-------|-------|-------|
| AU | 16.68 | 16.61 | 16.54 | **14.98** | 14.64 | 14.63 | 14.62 | **14.44** |
| DI | 13.80 | 13.79 | 13.77 | **13.65** | 13.25 | 13.25 | 13.24 | **13.23** |
| GL | 14.22 | 14.19 | 14.13 | **12.14** | 11.38 | 11.34 | 11.30 | **10.25** |
| HD | 12.99 | 12.95 | 12.90 | **12.41** | 11.75 | 11.74 | 11.73 | **11.67** |
| IR | 4.34  | 4.22  | 4.08  | **3.60**  | 3.20  | 3.17  | 3.14  | **3.07**  |
| LY | 19.27 | 19.27 | 19.25 | **16.78** | 15.85 | 15.89 | 15.90 | **14.73** |

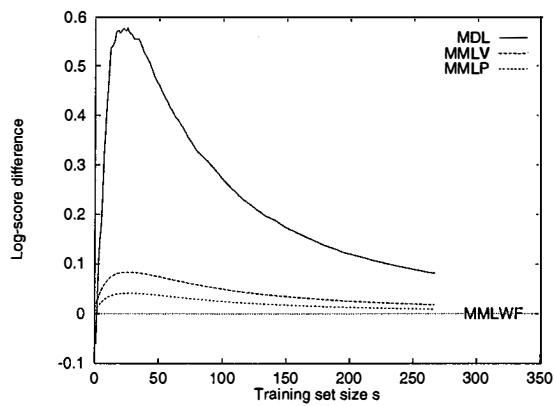

Figure 1: The HD dataset leave-one-out crossvalidated results as a function of the training data available, scaled with respect to the MMLWF score. The higher the score, the better the prediction accuracy.

Looking at Figure 1 again, we hypothesize that for extremely small sample sizes, our asymptotic results simply do not apply; then, as more data arrives, we enter a region where they do apply and the performance of the three MML methods is as predicted by the theory. Then, as the sample size grows truly large, the law of large numbers 'takes over' and the differences between the three methods become negligible.

## 7 CONCLUSION

We have shown that the claimed similarity between Wallace and Freeman's MML approach and Rissanen's 1996 MDL approach is superficial, and that when applying the approaches for predictive modeling, we arrive at quite different methods in practice. Furthermore, we pointed out how a technical weakness in the derivation of the MMLWF estimator can be corrected, and introduced two revised versions of the MML estimator, of which the volumewise optimal MMLV estimator was shown to be related to Rissanen's MDL estimator.

In order to apply the theoretical constructs for predictive modeling purposes, we showed how to develop different prediction methods based on the the minimum encoding estimators presented. As the theoretical results presented here are asymptotic in nature, this raised the question of the small sample behavior of these methods. To be able to study this question empirically, we instantiated the different prediction methods in the Bayesian network model family case.

In the empirical tests performed, it was observed that while in simple classification tasks the methods showed quite similar performance, in joint probability distribution estimation the MDL approach produced consistently the best results. What is more, the revised MML estimators introduced here gave usually better results than the MMLWF estimator. The fact that MDL performed better is probably largely due to the fact that the MDL approach used here is based on integrating over models instead of predicting using a single model; the fact that revised MML estimators work slightly better than the original MML estimators supports our theoretical analysis. The differences were larger with small amount of training data, and the differences between the various approaches became smaller with increasing amount of data, as was also expected from the theory.

**Acknowledgements.** This research has been supported by the ESPRIT Working Group on Neural and Computational Learning (NeuroCOLT), the Technology Development Center (TEKES), and the Academy of Finland.

# References

Baxter, R., & Oliver, J. (1994). *MDL and MML: Similarities and differences* (Tech. Rep. No. 207). Department of Computer Science, Monash University.

Berger, J. (1985). *Statistical decision theory and Bayesian analysis.* New York: Springer-Verlag.

Cooper, G., & Herskovits, E. (1992). A Bayesian method for the induction of probabilistic networks from data. *Machine Learning, 9,* 309–347.

Cover, T., & Thomas, J. (1991). *Elements of information theory.* New York, NY: John Wiley & Sons.

Friedman, N., Geiger, D., & Goldszmidt, M. (1997). Bayesian network classifiers. *Machine Learning, 29,* 131–163.

Heckerman, D. (1996). *A tutorial on learning with Bayesian networks* (Tech. Rep. No. MSR-TR-95-06). One Microsoft Way, Redmond, WA 98052: Microsoft Research, Advanced Technology Division.

Heckerman, D., Geiger, D., & Chickering, D. (1995). Learning Bayesian networks: The combination of knowledge and statistical data. *Machine Learning, 20*(3), 197–243.

Kontkanen, P., Myllymäki, P., Silander, T., Tirri, H., & Grünwald, P. (1997). Comparing predictive inference methods for discrete domains. In *Proceedings of the sixth international workshop on artificial intelligence and statistics* (pp. 311–318). Ft. Lauderdale, Florida.

Kontkanen, P., Myllymäki, P., Silander, T., Tirri, H., & Grünwald, P. (1998). Bayesian and information-theoretic priors for Bayesian network parameters. In C. Nédellec & C. Rouveirol (Eds.), *Machine learning: ECML-98, proceedings of the 10th European conference* (pp. 89–94). Springer-Verlag.

Rissanen, J. (1978). Modeling by shortest data description. *Automatica, 14,* 445–471.

Rissanen, J. (1987). Stochastic complexity. *Journal of the Royal Statistical Society, 49*(3), 223–239 and 252–265.

Rissanen, J. (1989). *Stochastic complexity in statistical inquiry.* New Jersey: World Scientific Publishing Company.

Rissanen, J. (1996). Fisher information and stochastic complexity. *IEEE Transactions on Information Theory, 42*(1), 40–47.

Schwarz, G. (1978). Estimating the dimension of a model. *Annals of Statistics, 6,* 461–464.

Wallace, C., & Boulton, D. (1968). An information measure for classification. *Computer Journal, 11,* 185–194.

Wallace, C., & Freeman, P. (1987). Estimation and inference by compact coding. *Journal of the Royal Statistical Society, 49*(3), 240–265.